\documentclass[logo, address]{trfr} 

\usepackage{hyperref}
\usepackage{amsmath}
\usepackage{amssymb}

\usepackage{booktabs}  

\usepackage{multirow}
\usepackage{makecell}
\usepackage{pifont}
\newcommand{\cmark}{\ding{51}}
\newcommand{\xmark}{\ding{55}}
\usepackage{xcolor}
\usepackage{wrapfig}
\usepackage{graphicx}
\usepackage{csquotes}

\usepackage[
  backend=biber,
  natbib=true,
  sorting=none,
  style=numeric,
  maxnames=5,
  minnames=1
]{biblatex}
\title{Scales++: Compute Efficient Evaluation Subset Selection with Cognitive Scales Embeddings}

\shorttitle{Scales++: Compute Efficient Evaluation Subset Selection with Cognitive Scales Embeddings}

\abstract{
The prohibitive cost of evaluating large language models (LLMs) on comprehensive benchmarks necessitates the creation of small yet representative data subsets (i.e., tiny benchmarks) that enable efficient assessment while retaining predictive fidelity. Current methods for this task operate under a model-centric paradigm, selecting benchmarking items based on the collective performance of existing models. Such approaches are limited by large upfront costs, an inability to immediately handle new benchmarks (`cold-start'), and the fragile assumption that future models will share the failure patterns of their predecessors. In this work, we propose a new item-centric approach to benchmark subset selection, arguing that selection should be based on the intrinsic properties of the task items themselves, rather than on model-specific failure patterns. We instantiate this item-centric efficient benchmarking approach via a novel method, \textsc{Scales++}, where data selection is based on the cognitive demands of the benchmark samples. Empirically, we show \textsc{Scales++} reduces the upfront selection cost by over 18$\times$ while achieving competitive predictive fidelity. On the Open LLM Leaderboard, using just a 0.25\% data subset, we predict full benchmark scores with a 3.2\% mean absolute error, and on Humanity's Last Exam we predict full scores with 2.9\% mean absolute error using a 2.0\% sample. We demonstrate that this item-centric approach enables more efficient model evaluation without significant fidelity degradation, while also providing better cold-start performance and more interpretable benchmarking.
}

\author[1,2]{Andrew M. Bean}
\author[1]{Nabeel Seedat}
\author[1,3]{Shengzhuang Chen}
\author[1,3]{Jonathan Richard Schwarz}

\affiliation[1]{Thomson Reuters Foundational Research}
\affiliation[2]{University of Oxford}
\affiliation[3]{Imperial College London}

\correspondence{\{first.last\}@thomsonreuters.com}

\begin{document}

\maketitle

\section{Introduction}

Large language models (LLMs) have demonstrated the ability to perform well on a broad range of tasks, including adapting to new tasks with little or no additional training \citep{brownLanguageModelsAre2020}. Consequently, evaluating LLMs across broad benchmark suites has become central to tracking progress, guiding training, and informing deployment \citep{rajiAIEverythingWhole2021}. However, running full evaluations is increasingly expensive in terms of energy and compute resources as models and datasets scale \citep{kaplanScalingLawsNeural2020, liang2023holistic, hendrycksMeasuringMassiveMultitask2021}, and repeated re-evaluation during model development cycles exacerbates this cost, which can run into \$10,000s for large or token-intensive (e.g. heavy CoT) benchmarks \cite{perlitz2024efficientbenchmarkinglanguagemodels}.

To address this, recent work has focused on performing evaluation using small, carefully selected data subsets that can reliably predict a model's performance on the full dataset with high fidelity. Most existing approaches to selecting this subset of benchmark examples for scoring are \textbf{model-centric}: they construct the subset by exploiting similarities in \emph{past model behaviour}. e.g. by clustering items using cross-model prediction statistics  \citep{vivekAnchorPointsBenchmarking2024a}  or by fitting psychometric Item Response Theory (IRT) parameters from historical item-level outcomes \citep{poloTinyBenchmarksEvaluatingLLMs2024a}. 

This assumption of access to item-level predictions of previous models on the target benchmark creates the following challenges: (i) It front-loads curation cost into running many models over many items (ii) It fails in \emph{cold-start} regimes where comparable logs are unavailable (new benchmarks/private model families) and (iii) It can struggle to \emph{transfer} when behaviour learned from one family does not generalize to another.

\begin{table*}[t]
\centering
\caption{Comparison of efficient LLM evaluation methods. Model-centric approaches require extensive upfront cost due to the need for historical model evaluations, before evaluating new models. In contrast, our item-centric $\textsc{Scales++}$ achieves an 18$\times$ reduction in setup cost while uniquely enabling cold-start evaluation through cognitive demand annotations rather than prior model performance patterns.}
\label{tab:related-work}
\resizebox{\textwidth}{!}{%
\begin{tabular}{lllcccc}
\toprule
\textbf{Method} & \textbf{Paradigm} & \textbf{Core Assumption} & \textbf{Upfront Cost} & \makecell{\textbf{No Historical}\\\textbf{Data Needed}} & \makecell{\textbf{Cold-Start}\\\textbf{Evaluation}} & \textbf{Interpretable} \\ 
\midrule
\makecell[l]{Anchor Points\\\citep{vivekAnchorPointsBenchmarking2024a}} & Model-centric & \makecell[l]{Past model correlations\\predict future correlations} & \makecell{$N$ models on\\full dataset\\($N \geq 10$)} & \xmark & \xmark & \xmark \\
\midrule
\makecell[l]{tinyBenchmarks\\\citep{poloTinyBenchmarksEvaluatingLLMs2024a}} & Model-centric & \makecell[l]{Past model failure patterns\\predict future patterns} & \makecell{319 models on\\full dataset} & \xmark & \xmark & \xmark \\
\midrule
\makecell[l]{metaBench\\\citep{kipnisMetabenchSparseBenchmark2025a}} & Model-centric & \makecell[l]{More past models $\rightarrow$\\better future prediction} & \makecell{5000+ models on\\full dataset} & \xmark & \xmark & \xmark \\
\midrule
\textbf{$\textsc{Scales++}$ (Ours)} & \textbf{Item-centric} & \makecell[l]{\textbf{Cognitive demands of task items}\\\textbf{predict performance}} & \makecell{\textbf{16 annotations}\\\textbf{per item}\\\textbf{(1 with GNN)}} & \cmark & \cmark & \cmark \\
\bottomrule
\end{tabular}%
}
\end{table*}

In this work, we challenge this dominant paradigm and propose an \textbf{item-centric} approach to benchmark subset selection. We argue that selection should be guided by the intrinsic properties of the task items themselves, rather than by model-specific failure patterns. We instantiate this approach with \textsc{Scales++}, a novel method that captures the intrinsic cognitive demands of benchmark samples. Drawing inspiration from the General Scales framework \citep{zhouGeneralScalesUnlock2025a}, which defines cognitive capabilities, we annotate each item along 16 cognitively grounded dimensions (e.g., logical reasoning, specific knowledge areas), yielding embeddings of item demands. We then (i) select a small, diverse subset in this space and (ii) predict full-benchmark performance via a combination of cluster-weighted estimates and per-dimension predictors, \emph{without the need for any historical data}.

To amortize annotation costs across datasets, we distill General Scales using a lightweight Graph Neural Network (GNN) predictor trained on a small auxiliary dataset with ground-truth GPT-4o annotations. This predictor leverages frozen embeddings from a pre-trained, open-source LLM and requires only a single forward pass for scales prediction per benchmark instance, dramatically reducing upfront annotation costs. We term this approach \textsc{Scales++ Lite}.

Our item-centric approach successfully addresses the limitations of prior work while maintaining competitive performance. Empirically, we demonstrate that \textsc{Scales++} reduces the upfront selection cost by over 18X while achieving high predictive fidelity. On the Open LLM Leaderboard, using just a 0.25\% data subset, we predict full benchmark scores with a 3.2\% mean absolute error; Our \textsc{Scales++ Lite} annotates the entire leaderboard in under 20 minutes, while outperforming expensive IRT baselines that require 300x more LLM calls for subsets of 0.5\% and 1.0\% of the evaluation data. We make three main contributions:

\textbf{Contributions:}\\
\ding{172}  We introduce a new \emph{item-centric paradigm} for benchmark subset selection that decouples selection from historical performance, thereby overcoming the high upfront costs and cold-start limitations of existing model-centric methods.

\ding{173}  We present \textsc{Scales++}, a novel item-centric method that creates interpretable embeddings based on the cognitive demands of task items, moving beyond a reliance on model failure patterns. We further introduce \textsc{Scales++ Lite}, a GNN-based predictor that allowing us to reduce per-item annotation costs for new datasets.

\ding{174} We show on the Open LLM Leaderboard’s six tasks and Humanity's Last Exam that the Scales variants match or surpass model-centric baselines while cutting up-front costs by over 18x, enabling cold-start evaluation without historical item-level logs and cost-effective re-evaluation during model development cycles.

\ding{175} We release a \textsc{tinyHLE} subset of the Humanity's Last Exam benchmark to be used for efficient benchmarking, along with our embedding annotations.

\section{Related Work}

This work engages with works on efficient evaluation and cognitive science in relation to LLMs. A contrast of key related work is presented in Table \ref{tab:related-work}.

\subsection{Efficient evaluation}
The escalating computational costs of evaluating increasingly large language models have motivated substantial research into efficient evaluation methodologies. Multiple studies have established that significant redundancy exists across benchmark items, with \cite{ye2023how} proposing to reduce the number of items in \texttt{Big-bench}~\citep{srivastava2023beyond}, while \cite{perlitz2024efficientbenchmarkinglanguagemodels} demonstrated that evaluation on \texttt{HELM}~\citep{liang2023holistic} relies on diversity across datasets but employs an excessive number of evaluation examples/items. \par

Building on these insights, benchmark curation and evaluation data selection methods have emerged as viable strategies for maintaining evaluation quality while reducing computational cost. \cite{li2025from} introduced the \texttt{BenchBuilder} pipeline, which leverages LLMs to curate high-quality prompts from large, crowd-sourced datasets through automated filtering based on seven quality indicators. Their approach was used to create \texttt{Arena-Hard-Auto}, a curated 500-item benchmark that capable of robustly recovering LLM relative rankings across multiple large benchmarks. 

The closest to our work are recent efforts to perform evaluation using small, selected evaluation subsets that can predict a model's performance on the full benchmark. \cite{vivekAnchorPointsBenchmarking2024a} proposed the Anchor Points method for evaluation subset selection, which advocates for reducing evaluation examples while maintaining accurate performance assessments. \cite{poloTinyBenchmarksEvaluatingLLMs2024a} proposed the \texttt{tinyBenchmarks}, demonstrating that full performance can be reliably estimated on benchmarks such as \texttt{MMLU} and \texttt{HELM} within 2\% mean absolute error leveraging trained IRT models (Item Response Theory) on evaluation results of 319 existing trained models on a small carefully curated subset of evaluation data. 

Most recently, \citet{kipnisMetabenchSparseBenchmark2025a} introduced \texttt{metabench}, which compresses the entire \texttt{Open LLM Leaderboard}~\citep{open-llm-leaderboard}—a collection of LLM benchmarks—to less than 3\% of its original size, providing reliable performance prediction and latent skill assessment by leveraging fitted IRT models trained on evaluation results from $>$5,000 trained LLM models. As highlighted in Table \ref{tab:related-work}, a key challenge with these methods is the model-centric assumption that past model performance is helpful for selection. Consequently, these methods rely on historical data as the basis for selection, and hence have a significant upfront cost, before evaluating new models. 

We directly address this challenge with our item-centric \textsc{Scales++} approach which reduces the setup cost by 18x, while maintaining similar performance (see Sec. \ref{sec:scales++})

\subsection{Cognitive approaches}
Recent work has begun exploring cognitive demand analysis as a means to better understand what LLM benchmarks actually measure by understanding the underlying cognitive requirements of evaluation tasks\cite{kearns2026quantifyingconstructvaliditylarge}. This research direction seeks to decompose benchmark items into their constituent cognitive challenges, such as reasoning complexity, knowledge requirements, and processing demands, providing a more principled understanding of why certain tasks prove difficult for models. Initial efforts in this space had limited success, finding a single dominant factor that strongly correlated to model size \cite{ILIC2024101858}.\par
The General Scales framework~\citep{zhouGeneralScalesUnlock2025a} implements this cognitive demand approach to greater success, operationalizing concepts from cognitive science to systematically analyze AI evaluation tasks. This framework operates by evaluating task items across multiple carefully crafted rubrics that systematically assess cognitive demands on scales ranging from 0 to 5, encompassing core cognitive abilities, knowledge domains, and task-related factors drawn from established cognitive science frameworks such as the Cattell-Horn-Carroll structure of human cognitive abilities~\citep{mcgrew2005cattell}. The scales can be applied automatically using LLMs to annotate evaluation task items~(see example in Appendix~\ref{app:scales}), making the approach scalable to tag datasets. 


While cognitive demand analysis was originally developed to understand and interpret benchmarks, we recognize its potential for addressing the orthogonal problem of efficient evaluation. Consequently, our work builds upon this foundation by adapting the General Scales framework for benchmark subset selection. We leverage the cognitive demand characterization provided by the 16-dimensional scale embeddings to identify representative evaluation instances, representing a novel item-centric approach to the problem of efficient evaluation. By decomposing task items into their constituent cognitive demands, our method facilitates a more principled selection of evaluation items that preserve the cognitive diversity essential for model evaluation.

\section{Methods}

\subsection{Problem Setting}

We consider the task of selecting a subset of items from a benchmark that best allows us to predict the overall score of a model on the benchmark. In this setting, evaluation of the model is costly and therefore only permitted on the selected subset of items, but the overall prediction may use other properties of the remaining benchmark items. This setting is similar to previous works \citep{poloTinyBenchmarksEvaluatingLLMs2024a, kipnisMetabenchSparseBenchmark2025a}, but we do not assume free access to the scores of other models on the benchmark.

More formally, we consider the task of predicting the performance of a model $\phi_m$ on a benchmark $B = (\{t_i\}_{i=1}^{N},M)$ consisting of a set of items $\{t_i\}$ and a metric $M$ which assigns scores to each (model, item) pair, $M: \{\phi_m\}\times\{t_i\} \rightarrow [0,1]$. Evaluating $M(\phi_m, t_i)$ is costly, so we want to select $I_{\text{sub}} \subset \{t_i\}_{i=1}^N$ with $|I_{\text{sub}}| = k \ll N$ such that we can predict $\frac{1}{N}\sum_{i=1}^N M(\phi_m, t_i)$ without evaluating $\{M(\phi_m, t_i)\}_{t_i\notin I_{\text{sub}}}$. In this paper, we focus on the case where M gives a binary correct/incorrect score for each model generation, though in principle, the setup can be generalized to any metric.

We measure the cost of creating the overall predictions in terms of calls to an LLM, where, for simplicity, we count all LLMs as equivalent in cost. In our problem setting, evaluating a subset of the benchmark reduces the number of LLM calls from $N$ to $k$ per model, with additional upfront costs of $\ell N$ LLM calls to provide information about the benchmark items. The total cost of scoring $m$ models on the benchmark is then $\text{Cost}(m) = km + \ell N$,  making $\ell$ a significant factor in determining the total cost of the evaluations in most cases.

\subsection{Item Selection for Evaluation}

\subsubsection{The Model-Centric Selection Paradigm}

Existing approaches operate under a \textbf{model-centric} paradigm.  Such approaches assume access to historical, item-level behavior from a set of prior models, $\Phi = \{\phi_1, \phi_2, \dots, \phi_n\}$, captured in a performance matrix $Y$. Prior methods like Anchor Points \citep{vivekAnchorPointsBenchmarking2024a} and \texttt{tinyBenchmarks} \citep{poloTinyBenchmarksEvaluatingLLMs2024a} use this matrix to guide subset selection. The process typically involves two stages: embedding and selection.

\textbf{1. Item Embedding} Each item $t_i$ is mapped to a low-dimensional embedding, $E_i$, that is a function of the collective performance of the source models.
\begin{itemize}
    \item Direct Performance Embedding: The embedding for item $t_i$ is the vector of performance scores from all source models, i.e., the $i$-th column of $Y$. This is the basis for the Anchor Points method \citep{vivekAnchorPointsBenchmarking2024a}, which uses the correlation between these vectors to define a distance metric for clustering.
    
    \item IRT-based Embedding: An Item Response Theory (IRT) model is fit to the entire performance matrix $Y$. The learned IRT parameters for item $t_i$ (e.g., discrimination $\alpha_i$ and difficulty $\beta_i$) form its embedding $E_i$. The \texttt{tinyBenchmarks} method uses this approach \citep{poloTinyBenchmarksEvaluatingLLMs2024a}. The embedding $E_i$ is thus a function of the full matrix, $E_i = g_{\text{IRT}}(Y, i)$.
\end{itemize}

\textbf{2. Item Selection} A subset $I_{\text{sub}}$ is chosen by applying a clustering algorithm (e.g., K-Means or K-Medoids) to the set of all item embeddings $\{E_1, \ldots, E_N\}$.

Crucially, the entire selection process is a function of the performance matrix. i.e. $I_{\text{sub}} = f_{\text{mc}}(Y, k).$ This direct dependency on $Y$ leads to high upfront computation costs (to generate $Y$ by running $M$ source models on all $N$ items) and the inability to evaluate new models in cold-start scenarios where $Y$ is unavailable.

\subsubsection{The Item-Centric Selection Paradigm (Ours)}

To address these issues, we propose a \textbf{item-centric} paradigm that decouples the selection process from historical model performance. The selection function depends only on the intrinsic, observable properties of the task items themselves.
Specifically, we assume an item-centric paradigm has an \emph{intrinsic} feature map $\psi: \mathcal{T} \to \mathbb{R}^D$ that depends only on the content of a task item $t_i \in \mathcal{T}$, not on any model's performance on it.

\textbf{1. Item Embedding.} Each item $t_i$ is mapped to an embedding $C_i$ via a model-agnostic annotation function, $\psi$, that analyzes the content of the item. i.e. $C_i = \psi(t_i)$

In our work, \textsc{Scales++}, we instantiate $\psi$ as a \textbf{Cognitive Scales annotation process} building upon \citet{zhouGeneralScalesUnlock2025a}. This function maps each task item $t_i$ to a 16-dimensional vector $C_i \in \mathbb{R}^{16}$, where each dimension corresponds to a specific cognitive skill or knowledge domain (e.g., logical reasoning, knowledge of social sciences). This annotation is performed using an LLM (e.g., GPT-4o) applied to a static, pre-defined rubric, making it independent of any specific model's success or failure on the item.

\textbf{2. Item Selection.} As in the model-centric paradigm, a subset $I_{\text{sub}}$ is chosen by clustering the set of embeddings $\{C_1, \ldots, C_N\}$. In our implementation, we first use UMAP for dimensionality reduction before applying k-means clustering. The key distinction is that our selection process is a function of the task set $\mathcal{T}$, not the performance matrix $Y$. i.e. $I_{\text{sub}} = f_{\text{ic}}(\mathcal{T}, k).$

By removing the dependency on $Y$, the item-centric paradigm inherently resolves the cold-start problem and dramatically reduces the upfront cost of subset selection. The annotation cost is also model-independent and can be amortized across all future model evaluations. In addition, in Sec. \ref{sec:gnn}, we show how a distilled predictor can be used in place of the annotation function $\psi$, as an alternative model-agnostic annotation function.

\subsubsection{Baseline Subset Selection Methods}


\textbf{Random.} We randomly select $k$ items from the benchmark, evaluate the target model on these items, and compute an average of the scores as a prediction of the overall score. For multi-task benchmarks such as the Open LLM Leaderboard, we stratify the sampling to ensure proportional representation for each constituent benchmark. 

\textbf{Clustering.} The scores from evaluating separate LLMs are used as an embedding for each item. Using these embeddings, the `anchor points' are selected as the solution to a k-medoids problem~\citep{rdusseeun1987clustering}. This approach is based on the Anchor Points method introduced in \citet{vivekAnchorPointsBenchmarking2024a}, though we use the more recent implementation of \citet{poloTinyBenchmarksEvaluatingLLMs2024a}, which selects the points closest to the k-means centers \citep{mcqueen1967some} as the items to use for scoring. The overall score prediction is a weighted average of the score on these points, with weights proportional to the number of other points in the cluster corresponding to each anchor point.

\textbf{IRT.} Scores from separate LLM evaluations are used to fit an IRT model to the benchmark items. We use the hyperparameters from \citep{poloTinyBenchmarksEvaluatingLLMs2024a}, which introduced this approach, fitting a two-parameter 3-dimensional model. These parameters are then used as embeddings for the items in the benchmark, and points are selected with k-means clustering. The p-IRT estimator uses the learned IRT model with a weighted average of the scores on the cluster center items. For the gp-IRT estimator, this is combined with an estimator using a weighted average of the model scores on the cluster center items to form a final estimate. \footnote{Recent work in \citet{kipnisMetabenchSparseBenchmark2025a} (metaBench) has noted an IRT model using results from only about 300 other models is likely to be underfit. They address this by training on results from $>$5000 other models and show better results, but due to the impracticality of having access to 5000 other model runs in most cases, we do not include their approach as a baseline.}

\subsubsection{Scales++ (Ours)}\label{sec:scales++}

\begin{figure}
    \centering
    \includegraphics[width=0.6\linewidth]{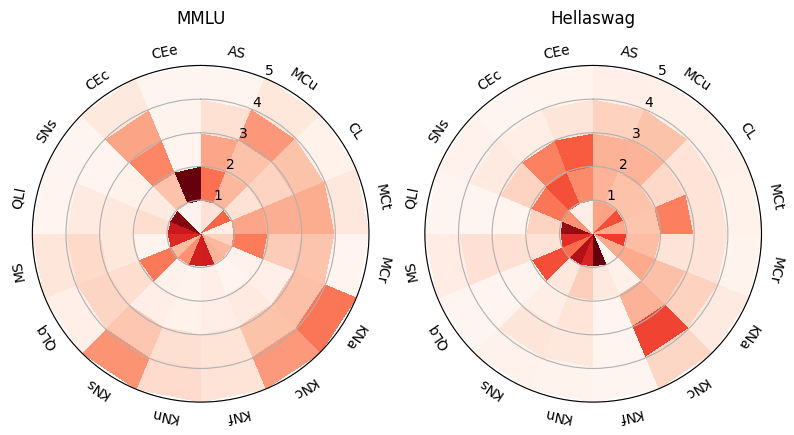}
    \includegraphics[width=0.35\linewidth]{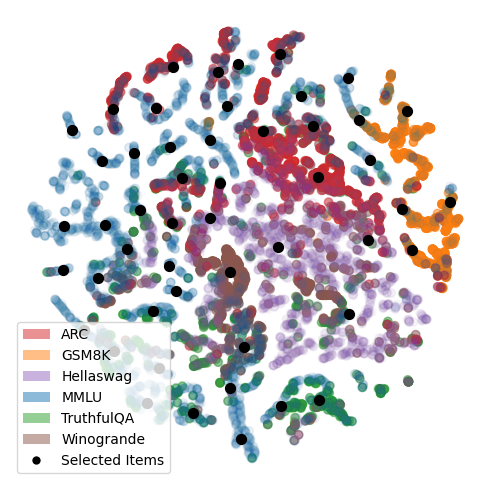}
    \caption{\textbf{Scales++ Item Selection.} For two example benchmarks, radial plots show the distribution of items which demand each level of capability along the 16 dimensions. MMLU often requires higher levels of subject area knowledge (KNa, KNc, KNs), but the two benchmarks have similar profiles along many dimensions, making the combined Open LLM Leaderboard a good candidate for subset selection. The embedded items are shown in a 2D t-SNE plot with the selected items highlighted in black, and the consituent benchmarks shown by the colors.
    }
    \label{fig:scales_selection}
\end{figure}

Our method begins by creating annotations for the degree to which each benchmark item requires 16 different cognitive skills (see Figure~\ref{fig:scales_selection}). These skills range from `logical reasoning' to `knowledge of social sciences', covering basic cognitive skills as well as knowledge in specific content areas. We use GPT-4o to annotate each item of the benchmark since \citet{zhouGeneralScalesUnlock2025a} provide expert-validated prompts optimized for this model. This creates a rating on a scale from 0-5 for each dimension.

We take these annotations as a 16-dimensional embedding of the benchmark items, which needs to be reduced to lower dimensions for effective clustering. We first discard any dimensions with no variation (This is possible if a benchmark has very similar items along a dimension. For example, a mathematics benchmark is unlikely to require any social science knowledge in any item.) In preparation for clustering, we then use UMAP \citep{mcinnes2018umap} to reduce the embeddings from 16 dimensions to 3 and then apply k-means clustering to select a subset of points. For the selected points, we evaluate the target LLM on each one, and estimate the overall benchmark score with a weighted average of the item scores weighted by the cluster sizes.

Leveraging the meaningful embedding dimensions, we fit a second estimator of item performance based on the difficulty levels of each item. We take the scores from the target model on each of the selected points from clustering and fit 16 separate logistic regressions for these points along each of the embedding dimensions. Based on the example of \citet{zhouGeneralScalesUnlock2025a}, we include additional data points with a performance of 0 at a difficulty of 20, representing a hypothetical maximum difficulty. For each remaining item in the benchmark, we predict the performance of the model using the average prediction of these regressions.

Our final estimator combines these two estimates with a weighted average. Specifically, our final estimator uses weight $\lambda = \hat{b}_2^2 / (\hat{b}_2^2 + \hat{v}_1)$ on the first, clustering-based estimator, and $(1-\lambda)$ on the second logistic regression estimator, where $\hat{b}_2$ is the estimate of the bias of the logistic regression estimator based on the selected items, and $\hat{v}_1$ is the estimate of the variance of the clustering-based estimator. These weights have been shown by \citet{songMinimal1988} to create optimal linear combinations of estimators based on their bias and variance. 

\begin{figure*}[!t]
    \centering
    \includegraphics[width=0.8\linewidth]{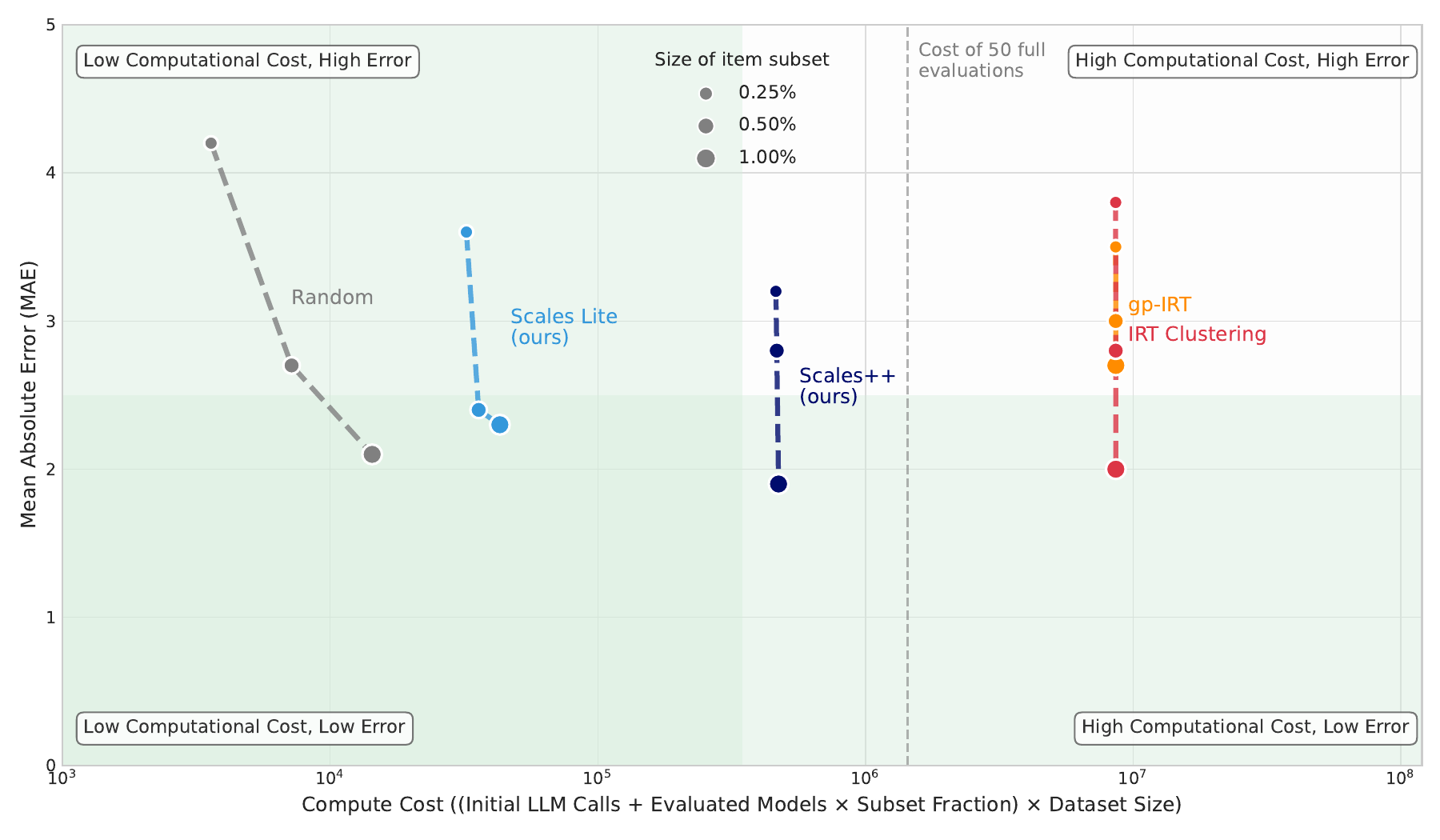}
    \caption{\textbf{Mean Absolute Estimation Error.} Performance estimation error in percentage points when selecting a subset of items to evaluate the target LLM and predict performance on the entire Open LLM Leaderboard. Marker size indicates the percentage of the benchmark being selected. For purposes of visualization, we have estimated the compute costs assuming that 50 evaluation runs will be performed using the subsampling methods. The p-IRT method is omitted due to extremely high MAE, but is listed in Appendix~\ref{app:A}. The \textsc{Scales} variants (\textsc{Scales++} and \textsc{Scales++ lite}) deliver the most favourable compute-performance trade-off across 3 out of 4 evaluation subset sizes.
    }
    \label{fig:main_results}
\end{figure*}

\subsection{Scalable GNN-Based Predictor for Cognitive Scales Embedding}\label{sec:gnn}

While our method demonstrates superior efficiency compared to prior approaches, the initial requirement of 16 GPT-4o calls per evaluation datapoint still incurs substantial computational costs. To further reduce this upfront expense, we train a lightweight neural network predictor that directly estimates the 16-dimensional cognitive scales embedding for any given task item, thereby eliminating the dependency on expensive GPT-4o inference.\par

The basic premise of our approach is to formulate the embedding prediction task as a standard supervised learning problem, where the objective is to replicate GPT-4o's cognitive assessment capabilities through a more computationally efficient model architecture\footnote{A natural approach would be to fine-tune a smaller LLM on GPT-4o outputs from identical query prompts used to obtain the scales embeddings. However, this approach still incurs significant computational cost from: (1) fine-tuning an LLM, and (2) autoregressive generation to label each point.}. To do this, we created a small set of auxiliary training data, consisting of 8,000 randomly subsampled queries from the Tulu3-SFT-mixture dataset~\citep{lambert2025tulu}, and labelled it with ground-truth GPT-4o-generated cognitive scales embeddings. We train a lightweight classifier which leverages embeddings from a pre-trained LLM. Through empirical evaluation of various neural network architectures,~
we found that a Graph Neural Network~(GNN)-based predictor consistently achieves the best performance on our validation dataset. 

To create the GNN, we embed each evaluation instance by feeding its query prompt to the Qwen2.5-7B-Instruct model~\citep{qwen2025qwen25technicalreport}, from which we extract token embeddings at the 14th layer (middle layer) and apply mean pooling to obtain a fixed-dimensional representation for each sample. These LLM representations serve as node features in our graph construction, where we formulate the prediction task as node classification with 16-dimensional labels corresponding to cognitive scale dimensions, each ranging from 0 to 5. To construct the edges of the input graph, we connect each node and its top-10 nearest neighbours based on cosine similarity in the embedding space. The trainable GNN classifier comprises three stacked graph convolutional layers~\citep{kipf2017semisupervised} and is optimized using cross-entropy loss. Model selection is determined by validation performance on a held-out split of our auxiliary training data. This design offers significant computational advantages at both training and inference stages: LLM embeddings can be readily extracted from open-source models without expensive API calls, prediction for any task item requires only a single, non-autoregressive, forward pass through the LLM and classifier network, and the upfront training cost of this predictor can be amortized across multiple benchmark evaluations, making it increasingly cost-effective with additional evaluation samples. Remarkably, \textsc{Scales++ Lite} can annotate the entire Open LLM Leaderboard, including 28,659 evaluation instances, in under 20 minutes, significantly reducing the computational requirements for obtaining cognitive scale embeddings while maintaining prediction quality, enabling our benchmark subset selection method to scale to larger evaluation datasets without prohibitive costs.


\section{Experiments}

We empirically validate the effectiveness of the \textbf{item-centric} paradigm (specifically \textsc{Scales++}) against the prevailing \textbf{model-centric} approaches. We also include a random selection baseline.

\textbf{Experimental Setup.} \label{sec:setup} We conduct our evaluation on the Open LLM Leaderboard \citep{open-llm-leaderboard}, a comprehensive suite comprising six diverse benchmarks: GSM8K \citep{cobbeTrainingVerifiersSolve2021} (1319 items), MMLU \citep{hendrycksMeasuringMassiveMultitask2021} (14042 items), Winograde \citep{sakaguchiWinoGrandeAdversarialWinograd2019} (1267 items), TruthfulQA \citep{linTruthfulQAMeasuringHow2022} (817 items), Hellaswag \citep{zellersHellaSwagCanMachine2019} (10042 items), and ARC \citep{clarkThinkYouHave2018} (1172 items). Collectively, this dataset contains 28,659 items, making it a prime candidate for down-sampling. While our method does \textit{not} require historical model data, we utilize the publicly released scores of 395 models \citep{poloTinyBenchmarksEvaluatingLLMs2024a} to train the baseline IRT/Clustering methods and to serve as ground truth for evaluating predictive error. 

We consider three subset budgets: 0.25\%, 0.5\%, and 1.0\% of the full leaderboard, corresponding to 71-287 items, similar to previous works testing samples between 50-300 items. As a heuristic, the 50-300 range is typically optimal for subsampling since smaller samples can be noisy and larger ones begin to saturate. Unless stated otherwise, we treat the 95 most recently released models as a held-out test set and train model-centric baselines on the remaining 300 models. We report mean absolute error (MAE; percentage points, $\downarrow$) between the estimated score from the subset and the true score computed on the full benchmark. For each method and budget, we run 10 repetitions with different random seeds to quantify variability from non-deterministic components (e.g., k-means, UMAP, IRT initialization).

\subsection*{Performance on the Open LLM Leaderboard}

Figure~\ref{fig:main_results} shows a comparison of \textsc{Scales++} and \textsc{Scales++ Lite} to baseline methods, both indicating compute cost and error prediction. We show that \textsc{Scales++} achieves \textbf{\textit{3.2\% MAE}} when sampling just 0.25\% of the benchmark (286 items), outperforming random selection by 1.0\% MAE and gp-IRT by 0.3\% MAE while requiring \textbf{\textit{95\% fewer initial LLM calls}} and avoiding the substantial historical-evaluation requirements of model-centric methods. Either \textsc{Scales++} or \textsc{Scales++ Lite} is the best performing method at each point, offering a viable "cold-start" solution where historical model logs are unavailable for IRT training. The random baseline is more competitive on larger samples, indicating that benchmark subsampling methods are most valuable in cases such as model development cycles, which involve many repeated evaluations across checkpoints so that the upfront costs of subset selection can be amortized.

We conduct the same testing for each of the constituent benchmarks of the Open LLM Leaderboard. For each benchmark, we select subsets of 1.0\% and 2.0\% and report the mean MAE on the held out models.
We omit the 0.25\% and 0.5\% subsets due to the small size of the individual benchmarks, meaning that most samples would have fewer than 10 items.

On individual benchmarks, results are more mixed, with different methods between \textsc{Scales++}, \textsc{IRT Clustering}, \textsc{gp-IRT} and random being best in different cases. \textsc{Scales++} is similar (within 2\% MAE) or better than all other methods in roughly 70\% of cases without requiring any prior model runs and requiring 95\% fewer LLM calls to create. TruthfulQA is particularly challenging for the Scales++ method, but this may be somewhat due to the noise when selecting just 16 items to represent the benchmark, as the variation across runs is high and Scales++ Lite is the best approach for 1.0\% and 2.0\% samples. We observe a performance advantage for \textsc{Scales++} relative to IRT methods on GSM8K. We compare the performance of the subset selection to the high semantic density of the dataset; only 52.3\% of benchmark items on GSM8K possess unique cognitive embeddings, compared to $\geq84\%$ for the other benchmarks, and as high as 96.6\% for TruthfulQA. This suggests that cognitive embedding density could serve as a meta-property for predicting which benchmarks benefit most from \textsc{Scales++}. Benchmarks with natural clustering in cognitive space (moderate uniqueness 60-85\%) offer the best balance: sufficient diversity for comprehensive evaluation, yet enough structure for representative sampling. Full results are included in Appendix~\ref{app:scales}.

\subsection{Cold-start Generalization on Humanity's Last Exam}

In order to test the generalization of the \textsc{Scales++} approach, we ran the subsetting pipeline on the Humanity's Last Exam benchmark (HLE) \cite{phanBenchmarkExpertlevelAcademic2026}. HLE is a large, diverse benchmark of extremely difficult question-answering tasks, making it a good candidate for subsetting. Unlike the Open LLM Leaderboard, a dataset of existing item-level evaluation results is not publicly available. This highlights the need for the cold start capabilities of our method, which can still be applied to HLE when IRT-based methods cannot. In order to create a small-scale test set, we ran full evaluations on a varied set of frontier-level models: GPT-5, Claude Sonnet 4.6, Claude Opus 4.7, Qwen 3 Next 80B A3B, DeepSeek v3.2, and Mistral Large 3, with inference details in Appendix~\ref{app:individual}.

Table~\ref{tab:hle} shows the MAE of the predicted scores of the models averaged across ten random seeds for samples of 0.5-2.0\% (Sample percentages were increased due to the smaller size of the benchmark). The Scales++ method consistently outperforms the random baseline, showing the ability of the \textsc{Scales++} method to generalize beyond the Open LLM Leaderboard and even to tasks outside the capabilities of the annotating model.

To support the community, we release a \textsc{tinyHLE} dataset consisting of a list of the best performing predictor items at each sample size and weights corresponding to the size of the clusters around each point. Our code implementation allows for the running of new models through the whole pipeline, but we also report the slightly higher MAE values of the clustering-based estimator alone as a baseline comparison to the combined estimator.

\begin{table}
\renewcommand{\arraystretch}{1.2}
  \centering
  \caption{\textbf{Mean Absolute Estimation Error: HLE.} Performance estimation error in percentage points for each benchmark. Values are reported as means over ten samples and the standard error of the means.}
  \begin{tabular}{cccc}
    \textbf{Subset} & Random & Scales++ & Scales-Clustering Only\\
     \hline
     0.5\% & 6.6 \scriptsize{± 0.3} & \textbf{5.6 \scriptsize{± 0.6}} & 6.3 \scriptsize{± 0.7} \\
     1\% & 5.3 \scriptsize{± 0.5} & \textbf{3.5 \scriptsize{± 0.3}} & 4.2 \scriptsize{± 0.3} \\
     2\% & 3.2 \scriptsize{± 0.4} & \textbf{2.9 \scriptsize{± 0.3}} & 3.4 \scriptsize{± 0.2} \\
     \hline
     
  \end{tabular}
  \label{tab:hle}
  
\end{table}

\section{Discussion \& Conclusion}
As the rapid development of LLMs continues, the need for efficient and reliable evaluation methods becomes increasingly critical. This work introduces a shift in approach from model-centric to item-centric benchmark subset selection, addressing fundamental limitations in current efficient benchmarking approaches. \textsc{Scales++} provides the best practical efficiency-accuracy trade-off, achieving 3.2\% error with only 0.25\% of items and minimal initialization cost. The comparison between 16 vs 300 LLM calls for initialization reveals IRT's hidden computational cost, making our approach 18X more efficient for comparable accuracy. This makes \textsc{Scales++} particularly valuable when evaluating on new benchmarks or working under computational constraints, as it translates to concrete benefits: \emph{a large model can be benchmarked in hours rather than days}. The \textsc{Scales++ Lite} variant reduces annotation costs further through our GNN-based predictor, enabling cheap benchmark annotation while maintaining competitive performance. Furthermore, we release the \textsc{tinyHLE} dataset, a subset of the HLE benchmark created via Scales++ which can predict performance on the overall benchmark with 2.9\% MAE from only a 2\% sample.

\label{sec:limitations}
We discuss the applicability and trade-offs of our approach. While performance is generally competitive or better than existing methods, significant advantages of our framework derive from its independence from historical model priors and its reduced cost. Consequently, we propose a clear heuristic for adoption: \textsc{Scales++} is best suited for cold-start scenarios (such as new/private benchmarks or novel architectures) and repeated evaluation scenarios (such as model training). Conversely, in situations where sufficient historical data is already available, model-centric methods remain strong candidates and could be considered. Our testing focused on the Open LLM Leaderboard and HLE, which have diverse task items, which means the Scales embeddings will have greater dimensional variation. In single domain settings, the embeddings may be less useful, though our results on individual benchmarks suggest that embedding density, rather than domain type alone, may be the more robust indicator of where \textsc{Scales++} is most effective. Finally, we note that all tested methods exhibit relatively high variance between random seeds, a challenge that future work could address by integrating these selection strategies with adaptive testing frameworks, such as in the recent work of \citet{hofmannFluidLanguageModel2025}.

Our work demonstrates that focusing on intrinsic task properties rather than historical model behaviour offers an efficient and potentially more generalizable path forward for comprehensive LLM assessment. The ability to achieve competitive mean absolute prediction error without training on the results of previous models represents a key new direction for approaching the increasingly important challenge of LLM evaluation at scale.

\clearpage
\printbibliography

\clearpage

\appendix
\section{Ablations and Additional Experiments}\label{app:A}

\subsection{Clustering Methods}

Clustering algorithms are used in combination with the task representations to select the final subset of tasks. The goal of this step is to select the points which best represent the benchmark as a whole. If the ability of a task to represent another task is measured by distance in the embedding space, then this problem is equivalent to a k-medoids problem, finding the set of $k$ points which minimizes the average distances to all of the other points in the dataset. For efficiency, we approximate the solution with the task item closest to the k-means centers, potentially affecting downstream performance. 

In this appendix, we validate the quality of the k-means approximation and compare to alternative clustering methods. We compare \textsc{K-means}, \textsc{K-medoids}, and \textsc{GMM} for selecting a subset of points with both IRT- and Scale-based embeddings. Since we are focusing on the clustering methods, we do not report results for the random baseline. We report the mean MAE and standard error across the $0.25\%, 0.5\%$ and $1.0\%$ subsets.

\begin{table*}[!h]
\centering
\small
\caption{Clustering method comparison (mean MAE ± standard error).}
\label{tab:exp-Scales}
\begin{tabular}{lccc}
\toprule
Method & \textsc{Scales++} & \textsc{Scales-Clustering Only}  \\
\midrule
\textsc{K-means}   & 4.1 \scriptsize{± 0.3} & 4.0 \scriptsize{± 0.4}\\
\textsc{K-medoids} & 3.3 \scriptsize{± 0.1} & 3.7 \scriptsize{± 0.1} \\
\textsc{GMM}       & 5.1 \scriptsize{± 0.2} & 5.3 \scriptsize{± 0.3} \\
\bottomrule
\end{tabular}
\end{table*}

Table \ref{tab:exp-Scales} shows that the theoretically-motivated K-mediods does achieve the lowest average MAE, but that k-means is a reasonable approximation. GMM shows meaningfully higher error rates.

\subsection{Numerical Results for Figure~\ref{fig:main_results}}

We present full numerical results for Figure~\ref{fig:main_results}, in Table \ref{tab:performance_prediction} below.

\begin{table}[ht]  
    \centering  
    \renewcommand{\arraystretch}{1.3}  
     \caption{\textbf{Mean Absolute Estimation Error:} Performance estimation error in percentage points when selecting a subset of \textit{n} items to evaluate the target LLM and predict performance on the entire benchmark. Small numbers are the standard errors across 10 repetitions. The Open LLM Leaderboard contains 6 sub-benchmarks, and we allow the subset to be selected from any of the question items. We include p-IRT here but not in Figure~\ref{fig:main_results} due to scale. We also include a 2.0\% sample to show the saturation effects as the samples get larger than those typically used.
    }  
    \begin{tabular}{ll|c}  
        \toprule
        \textbf{Subset} & \textbf{Method} & \textbf{Open LLM Leaderboard (MAE $\downarrow$)} \\  
        \hline  
        \multirow{6}{*}{0.25\%}  
         & Random & 4.2 \scriptsize{± 0.5} \\  
         & IRT Clustering & 3.8 \scriptsize{± 0.2} \\  
         & p-IRT & 6.2 \scriptsize{± 0.7} \\  
         & gp-IRT & 3.5 \scriptsize{± 0.2} \\  
         & \textsc{Scales++} & \textbf{3.2} \scriptsize{± 0.1} \\  
         & \textsc{Scales++ Lite} & 3.6 \scriptsize{± 0.2} \\  
        \hline  
        \multirow{6}{*}{0.5\%}  
         & Random & 2.7 \scriptsize{± 0.2} \\  
         & IRT Clustering & 2.8 \scriptsize{± 0.0} \\  
         & p-IRT & 7.6 \scriptsize{± 0.8} \\  
         & gp-IRT & 3.0 \scriptsize{± 0.2} \\  
         & \textsc{Scales++} & 2.8 \scriptsize{± 0.2} \\  
         & \textsc{Scales++ Lite}& \textbf{2.4} \scriptsize{± 0.2} \\  
        \hline  
        \multirow{6}{*}{1.0\%}  
         & Random & 2.1 \scriptsize{± 0.3} \\  
         & IRT Clustering & 2.0 \scriptsize{± 0.0} \\  
         & p-IRT & 7.7 \scriptsize{± 0.8} \\  
         & gp-IRT & 2.7 \scriptsize{± 0.3} \\  
         & \textsc{Scales++} & \textbf{1.9} \scriptsize{± 0.2} \\  
         & \textsc{Scales++ Lite} & 2.3 \scriptsize{± 0.3} \\  
        \hline  
        \multirow{6}{*}{2.0\%}  
         & Random & \textbf{1.4} \scriptsize{± 0.1} \\  
         & IRT Clustering & 1.5 \scriptsize{± 0.0} \\  
         & p-IRT & 8.1 \scriptsize{± 0.8} \\  
         & gp-IRT & 2.5 \scriptsize{± 0.3} \\  
         & \textsc{Scales++} & \textbf{1.4} \scriptsize{± 0.1} \\  
         & \textsc{Scales++ Lite} & 1.5 \scriptsize{± 0.1} \\  
        \hline  
    \end{tabular}  
   
    \label{tab:performance_prediction}  
\end{table}

\subsection{Individual Benchmark Results}
\label{app:individual}

We show the full results on individual benchmarks  in Table~\ref{tab:performance_detailed}. 

\begin{table*}
\caption{\textbf{Mean Absolute Estimation Error: Individual Benchmarks} Performance estimation error in percentage points for each benchmark. Values are reported as means over ten samples, with the standard error across ten samples. Benchmark sizes are noted under their names, with some of the samples being extremely small at 1\%-2\%.}
\resizebox{\columnwidth}{!}{
    \centering
    \renewcommand{\arraystretch}{1.2}
    \begin{tabular}{ll|cccccc}
        \toprule
         & & \textbf{ARC} & \textbf{GSM8K} & \textbf{TruthfulQA} & \textbf{HellaSwag} & \textbf{MMLU} & \textbf{Winogrande} \\
         \textbf{Subset} & \textbf{Method} & (1172) & (1319) & (817) & (10042) & (14042) & (1267) \\
        \hline
        \multirow{6}{*}{1.0\%}  
         & Random & 13.4 \scriptsize{± 1.1} & \textbf{7.9} \scriptsize{± 0.4} & 13.1 \scriptsize{± 2.8} & 3.0 \scriptsize{± 0.4} & 3.0 \scriptsize{± 0.2} & 10.3 \scriptsize{± 1.0} \\  
         & IRT Clustering & 11.4 \scriptsize{± 0.5} & 9.1 \scriptsize{± 0.3} & 12.1 \scriptsize{± 0.6} & 2.5 \scriptsize{± 0.1} & \textbf{2.8} \scriptsize{± 0.0} & 10.5 \scriptsize{± 0.5} \\  
         & p-IRT & \textbf{9.1} \scriptsize{± 1.1} & 36.7 \scriptsize{± 4.2} & 10.6 \scriptsize{± 1.0} & 4.1 \scriptsize{± 0.4} & 9.3 \scriptsize{± 1.3} & 8.0 \scriptsize{± 1.1} \\  
         & gp-IRT & 9.4 \scriptsize{± 0.9} & 8.8 \scriptsize{± 0.3} & 10.5 \scriptsize{± 0.7} & \textbf{2.3} \scriptsize{± 0.1} & 2.9 \scriptsize{± 0.1} & \textbf{6.9} \scriptsize{± 0.6} \\
        & \textsc{Scales++} & 13.0 \scriptsize{± 1.3} & 8.4 \scriptsize{± 0.5} & 18.6 \scriptsize{± 1.1} & 3.0 \scriptsize{± 0.4} & 3.1 \scriptsize{± 0.1} & 8.6 \scriptsize{± 0.9} \\
        & \textsc{Scales++ Lite} & 10.8 \scriptsize{± 0.4} & 8.4 \scriptsize{± 0.5} & \textbf{9.8} \scriptsize{± 0.8} & 2.5 \scriptsize{± 0.3} & 3.7 \scriptsize{± 0.2} & 20.2 \scriptsize{± 0.0} \\
        \hline
        \multirow{6}{*}{2.0\%}
        & Random & 8.5 \scriptsize{± 1.2} & \textbf{5.5} \scriptsize{± 0.3} & 9.8 \scriptsize{± 1.2} & 1.9 \scriptsize{± 0.2} & 2.2 \scriptsize{± 0.2} & 6.8 \scriptsize{± 0.6} \\
        & IRT Clustering & 7.9 \scriptsize{± 0.3} & 7.7 \scriptsize{± 0.2} & 8.5 \scriptsize{± 0.5} & \textbf{1.8} \scriptsize{± 0.1} & \textbf{2.0} \scriptsize{± 0.1} & 7.5 \scriptsize{± 0.3} \\
        & p-IRT & 9.2 \scriptsize{± 1.0} & 37.6 \scriptsize{± 4.0} & 10.8 \scriptsize{± 1.0} & 4.1 \scriptsize{± 0.4} & 9.7 \scriptsize{± 1.2} & 8.0 \scriptsize{± 1.1} \\
        & gp-IRT & \textbf{6.7} \scriptsize{± 0.4} & 7.5 \scriptsize{± 0.2} & 7.4 \scriptsize{± 0.6} & 2.1 \scriptsize{± 0.2} & 2.4 \scriptsize{± 0.2} & \textbf{5.3} \scriptsize{± 0.5} \\
        & \textsc{Scales++} & 8.5 \scriptsize{± 0.9} & 6.1 \scriptsize{± 0.3} & 13.2 \scriptsize{± 2.3} & 2.3 \scriptsize{± 0.2} & 2.5 \scriptsize{± 0.2} & 6.3 \scriptsize{± 0.5} \\
        & \textsc{Scales++ Lite} & 9.2 \scriptsize{± 1.0} & 6.1 \scriptsize{± 0.3} & \textbf{5.9} \scriptsize{± 0.1} & 2.6 \scriptsize{± 0.3} & 2.4 \scriptsize{± 0.2} & 8.2 \scriptsize{± 0.6} \\  
        \hline  
    \end{tabular}

    \label{tab:performance_detailed}
}
\end{table*}

\subsection{Compute Details}
\label{sec:compute_resources}

Experiments were run on AWS EC2 instances with A/H100 GPUs used for training the IRT and GNN models. Computing a single Scales++ subset required about two minutes of CPU time, used primarily for UMAP dimensionality reduction. Annotating the benchmark items with General Scales labels required 16 GPT-4o model calls per item, run over the course of a week. Computing IRT subsets took about half an hour of GPU time to fit each IRT model, and about a minute to select the subsets.

For the HLE experiment, the model details are shown in Table~\ref{tab:inference}. All models were run with temperature 0 and a token limit of 16000 tokens per question.

\begin{table}[]
    \centering
    \renewcommand{\arraystretch}{1.2}
    \caption{Model Inference Details}
    \begin{tabular}{ccc}
        \textbf{Model} & \textbf{Provider} & \textbf{Instance Name} \\
        \hline
        GPT-5 & Azure & gpt-5 \\
        Claude Sonnet 4.6 & Bedrock & anthropic.claude-sonnet-4-6 \\
        Claude Opus 4.7 & Bedrock & anthropic.claude-opus-4-7 \\
        Mistral Large 3 & Bedrock & mistral.mistral-large-3-675b-instruct \\
        Qwen 3 Next & Bedrock & qwen.qwen3-next-80b-a3b \\
        DeepSeek v3.2 & Bedrock & deepseek.v3.2 \\
    \end{tabular}

    \label{tab:inference}
\end{table}
\clearpage
\clearpage
\section{General Scales}\label{app:scales}
General Scales~\citep{zhouGeneralScalesUnlock2025a} represents a comprehensive framework for AI evaluation that can explain what common AI benchmarks really measure, extract ability profiles of AI systems, and predict their performance for new task instances. The methodology builds on 18 newly-crafted rubrics that place instance demands on general scales that do not saturate, providing a standardized approach to assess cognitive and knowledge-based abilities across diverse AI evaluation tasks. The list of dimensions used in our works include:

\begin{itemize}
\item Attention and scan
\item Calibrating knowns and unknowns
\item Conceptualisation learning abstraction
\item Critical thinking processes
\item Identifying relevant information
\item Knowledge applied science
\item Knowledge customary
\item Knowledge formal science
\item Knowledge natural science
\item Knowledge social science
\item Logical reasoning
\item Mind modelling and social cognition
\item Quantitative reasoning
\item Spatial reasoning and navigation
\item Verbal comprehension
\item Verbal expression
\end{itemize}

These scales are obtained through an automatic annotation process using GPT-4o, with each task instance rated from 0 to 5 on each dimension based on detailed rubrics, indicating how much that ability contributes to successful task completion. \par

Below is one dimension-specific prompt template, where $\{\{$instance$\}\}$ is replaced with the prompt from the task instance in the evaluation benchmark.

\begin{tcolorbox}[
    colback=blue!5!white,
    colframe=blue!75!black,
    title=Prompt for Attention and Scan,
    fonttitle=\bfseries,
    boxrule=1pt,
    arc=3pt,
    breakable
]
QUERY: The following rubric describes six distinct levels of *Attention and Scan* required by different tasks\\
\\
\# Attention and Scan (AS)
\\
\\
This criterion assesses the level of attention and scan required to focus on or locate specific elements within a given stream of information or environment in the whole process of solving a task. During this process, there is the need to actively scan for or retrieve elements that meet predetermined criteria. The level represents the extent to which the task requires locating and focusing on specific target information, ranging from situations where the target is immediately obvious to those requiring sustained tracking of multiple targets among numerous distractors—any elements that are irrelevant to solve the task, such as visual objects, sounds, pieces of text, noise, or other stimuli, but compete for attention with the target information—in complex, dynamic environments. The challenge is not on determining what to look for but focusing the attention to find it within a larger context. This differs from tasks where there's a need to identify which pieces of information are relevant from a set already under consideration. While both processes may overlap in complex tasks like reading comprehension or image understanding, “attention and scan” specifically focuses on the deployment of attention during scan processes when solving the task, rather than the selection or evaluation of information.
\\
\\
\#\# Levels
\\
\\
\#\#\# Level 0: None\\
No attention or scan is required. The target information is immediately obvious or is the only information present.\\
**Examples:**\\
- "Given a single word input, determine if it starts with a capital letter."\\
- "Look at the only object in the centre of the white page and tell what colour it is."\\
- "Is Madrid the capital of Spain?"\\
\\
\#\#\# Level 1: Very low\\
Minimal attention or scanning is required. The target information is easily distinguishable with little to almost no distraction.\\
**Examples:**\\
- "Find the only blue car in a car park full of red cars."\\
- "Find the letter 'X' among a row of 'O's"\\
- "Spot the tall tree in a row of short bushes."\\
\\
\#\#\# Level 2: Low\\
Some attention or basic scanning is required. The target information is visible among a few distractors or in a small scan area.\\
**Examples:**\\
- "Find all the vowels in the following sentence: 'The quick brown fox jumps over the lazy dog.'"\\
- "Find who's wearing glasses in this photo of students at commencement, with 2 rows of 5 students each, all facing forward, taken by a professional photographer."\\
- "Who authored the Queensberry rules, which were published in 1867 for the sport of boxing? Choices: A. John Douglas  (in his late twenties) B. John Graham Chambers (in his mid-twenties) C. Marquess of Queensberry (in his early thirties) D. James Figg (in his forties)."\\
\\
\#\#\# Level 3: Intermediate\\
Moderate attention and scan are required. The target information is mixed with several distractors or spread over a fairly large scan area.\\
**Examples:**\\
- "Find everyone wearing glasses in this casual BBQ photo where 15 people are gathered around a table. Some are sitting, some standing, some looking at the camera while others are in conversation."\\
- "In a 5-page technical document about basic geometry, locate all explicit references to the Pythagorean theorem (a² + b² = c²), where the equation appears 5 times mixed among references to 15 other geometric formulas, with occasional inconsistent equation numbering but standard mathematical notation."\\
- "As we all know, the Queensberry Rules are a set of rules for boxing that govern both amateur and professional matches. Who authored the Queensberry rules, which were published in 1867 for the sport of boxing? Choices: A. John Douglas  (in his late twenties) B. John Graham Chambers (in his mid-twenties) C. Marquess of Queensberry (in his early thirties) D. James Figg (in his forties) E. James Zou (in his fifties) F. Lucy Grande (in her late twenties) G. Xiaoxiao Li (in her early forties) H. Enrique Garcia (in his late thirties)."\\
\\
\#\#\# Level 4: High\\
Sustained tracking of one or various targets is required. The target information is in an environment mixed with numerous distractors and changing conditions. Requires some continuous monitoring amid competing signals.\\
**Examples:**\\
- "Listening to a symphony, identify all instances where the clarinet plays in a minor key, even when it's not playing the main melody."\\
- "Track three orange spheres among twenty red spheres as they move randomly across a black screen (40 cm × 30 cm) at varying speeds (1-3 cm/s), with spheres frequently intersecting paths and maintaining a minimum separation distance of 2 cm. Each sphere is 1 cm in diameter."\\
- "In a real-time video feed of a busy airport, finding the locations of ten blue suitcases."\\
\\
\#\#\# Level 5+: Very High\\
Requires sustained attention and scan for simultaneous tracking of multiple targets across different domains or contexts, with continuous adaptation to fast-changing conditions. The target information is extremely difficult to distinguish from distractors or is hidden in a vast or constantly changing environment.\\
**Examples:**\\
- "While seated courtside at a professional basketball game, track two specific players throughout the entire game as they move at speeds up to 8m/s, frequently cluster with other players during rebounds, and weave through screens and defensive formations."\\
- "Monitor four simultaneous video feeds of a crowded airport terminal from different angles, detecting subtle security-relevant changes (e.g. brief interactions < 2 seconds, crowd flow changes, small object exchanges) across feeds."\\
- "While monitoring multiple simultaneous customer service chat conversations in different languages, identify instances where customers are expressing the same underlying technical issue, even though they're describing it using different metaphors, technical terms, or cultural references specific to their region."\\
\\
TASK INSTANCE: \{\{instance\}\}
\\
\\
INSTRUCTION: Score the level of *Attention and Scan* demanded by the given TASK INSTANCE using a discrete value from 0 to 5. Use CHAIN-OF-THOUGHTS REASONING to reason step by step before assigning the score. After the CHAIN-OF-THOUGHTS REASONING STEPS, conclude your assessment with the statement: "Thus, the level of *Attention and Scan* demanded by the given TASK INSTANCE is: SCORE", where 'SCORE' is the integer score you have determined.
\end{tcolorbox}

\end{document}